\documentclass[sigconf,nonacm]{acmart}

\AtBeginDocument{%
  \providecommand\BibTeX{{%
    \normalfont B\kern-0.5em{\scshape i\kern-0.25em b}\kern-0.8em\TeX}}}

\setcopyright{acmcopyright}
\copyrightyear{2018}
\acmYear{2018}
\acmDOI{10.1145/1122445.1122456}

\acmConference[nonacm]{Woodstock '18: ACM Symposium on Neural
  Gaze Detection}{June 03--05, 2018}{Woodstock, NY}
\acmBooktitle{Woodstock '18: ACM Symposium on Neural Gaze Detection,
  June 03--05, 2018, Woodstock, NY}
\acmPrice{15.00}
\acmISBN{978-1-4503-XXXX-X/18/06}

\usepackage{multirow}

\begin{document}

%%
%% The "title" command has an optional parameter,
%% allowing the author to define a "short title" to be used in page headers.
\title{Transportation Scenario Planning with Graph Neural Networks}

%%
%% The "author" command and its associated commands are used to define
%% the authors and their affiliations.
%% Of note is the shared affiliation of the first two authors, and the
%% "authornote" and "authornotemark" commands
%% used to denote shared contribution to the research.
\author{Ana Alice Peregrino}
\email{aapp@cin.ufpe.br}
\affiliation{%
  \institution{Universidade Federal de Pernambuco}
  \city{}
  \state{}
  \country{}
}

\author{Soham Pradhan}
%\authornotemark[1]
\email{spradh8@uic.edu}
\affiliation{%
  \institution{University of Illinois at Chicago}
  \city{}
  \state{}
  \country{}
}

\author{Zhicheng Liu}
%\authornotemark[1]
\email{zhichengliu@seu.edu.cn}
\affiliation{%
  \institution{Southeast University}
  \city{}
  \state{}
  \country{}
}

\author{Nivan Ferreira}
\email{nivan@cin.ufpe.br}
\affiliation{%
  \institution{Universidade Federal de Pernambuco}
  \city{}
  \state{}
  \country{}
}

\author{Fabio Miranda}
%\authornotemark[1]
\email{fabiom@uic.edu}
\affiliation{%
  \institution{University of Illinois at Chicago}
  \city{}
  \state{}
  \country{}
}

%%
%% By default, the full list of authors will be used in the page
%% headers. Often, this list is too long, and will overlap
%% other information printed in the page headers. This command allows
%% the author to define a more concise list
%% of authors' names for this purpose.
\renewcommand{\shortauthors}{Peregrino et al.}

\newcommand{\myparagraph}[1]{\vspace{0.05cm}\noindent \textbf{#1}}

%%
%% The abstract is a short summary of the work to be presented in the
%% article.
\begin{abstract}
Providing efficient human mobility services and infrastructure is one of the major concerns of most mid-sized to large cities around the world.
A proper understanding of the dynamics of commuting flows is, therefore, a requisite to better plan urban areas.
In this context, an important task is to study hypothetical scenarios in which possible future changes are evaluated.
For instance, how the increase in residential units or transportation modes in a neighborhood will change the commuting flows to or from that region?
In this paper, we propose to leverage GMEL, a recently introduced graph neural network model, to evaluate changes in commuting flows taking into account different land use and infrastructure scenarios.
We validate the usefulness of our methodology through real-world case studies set in two large cities in Brazil.
\end{abstract}

%%
%% The code below is generated by the tool at http://dl.acm.org/ccs.cfm.
%% Please copy and paste the code instead of the example below.
%%

%%
%% Keywords. The author(s) should pick words that accurately describe
%% the work being presented. Separate the keywords with commas.
\keywords{Urban planning, transportation, urban data, scenario planning, graph neural networks}

%% A "teaser" image appears between the author and affiliation
%% information and the body of the document, and typically spans the
%% page.
% \begin{teaserfigure}
%   \includegraphics[width=\textwidth]{sampleteaser}
%   \caption{Seattle Mariners at Spring Training, 2010.}
%   \Description{Enjoying the baseball game from the third-base
%   seats. Ichiro Suzuki preparing to bat.}
%   \label{fig:teaser}
% \end{teaserfigure}

%%
%% This command processes the author and affiliation and title
%% information and builds the first part of the formatted document.
\maketitle

\section{Introduction}
\label{sec:intro}

Cities are complex environments that house the majority of the world's population; today, 55\% of the world's population lives in urban areas, and this is expected to increase to 68\% by 2050~\cite{un}.
For this reason, an enormous problem faced by governments and urban planners is how to plan for this new surge of people while solving the already challenging scenarios of the present.
One of the issues present in nearly all mid-sized and large cities is human mobility.
In particular, every day millions of people commute from home to work and limitations in the transportation infrastructure cause not only people to waste their time, but also an increase in pollution and health problems.
A proper understanding of the dynamics of commuting flows in a city is, therefore, a requisite to better plan urban areas.
Urban planners, transportation specialists, and city agencies more often than not rely on precedent and data analyzed in isolation to make decisions that can impact or transform a city.
With the growing availability of urban data and advances in machine learning, there are new opportunities for data-driven solutions to better support the exploration of possible alternate urban scenarios.
Such information can be used, for instance, by urban planners to guide the development of new neighborhoods, or transportation specialists to direct the deployment of new transportation modes.
In this context, the goal of this paper is to study the use of a graph neural network-based commuting flow prediction model to assist experts in the identification of the effects of infrastructure, land use, and/or policy changes on commuting flows.
Understanding the commuting flow can help answer many what-if questions in the planning stage, such as \emph{``If a new high-rise building is planned for a region, to what regions would people commute to work?''}, and \emph{``How to modify the transportation infrastructure to improve commuting efficiency?''}
%
% The volume of commuting flows between areas in a city is affected by a number of urban factors, such as land use, presence of public transportation and bike lanes, socioeconomic indicators of the population, etc.
%
To enable data-driven scenario planning, we take the first steps in leveraging the Geo-contextual Multitask Embedding Learner (GMEL) model, previously proposed in Liu et al.~\cite{liu2020gmel}, as our base model for predicting commuting flows based on geographic information (e.g., infrastructure, land use, transportation).
Commuting flows are defined as flows between a workers' residence location and a workplace location.
While major cities have the resources to collect and process high-resolution land use data, other cities do not have such capabilities, especially in countries in the Global South.
To test the effectiveness of our methods in cities in developing countries, we focus our efforts on two cities in Brazil. Using urban data from these cities, we train the GMEL model to predict flows based on different infrastructure and land use scenarios.
Through a set of case studies, we show how scenario planning methods based on graph neural networks can reveal important information to transportation experts.

% and we train the model using urban data 

% Recife and Curitiba. We train the model using urban data from 

% . One of the challenges of applying this model in different cities is the lack of detailed land-use and infra-structure data as present in major American cities such as NYC. To evaluate the performance of the model with different types of input, we train the model using urban data from two cities in Brazil, Recife and Curitiba,
%
%  using it to investigate propose a model to predict commuting flows based on infra-structure and land-use related features, and also to evaluate how the model performs in scenario planning cases in Recife.

%

\vspace{-0.1cm}
\section{Related Work}
\label{sec:related}

% In this section, we briefly survey previous work related to commuting flow prediction, graph representation learning and graph neural networks, and alternate planning.

\subsection{Commuting flow prediction}
Classical trip distribution works follow a gravity model, first introduced in the 1940s~\cite{zipf_p1_1946}, and assumes that the trip volume is proportional to the product of population of origin and destination and is inversely proportional to the distance between origin and destination. Modern extensions of the gravity model take into account several factors, such as demographics and land use, to more accurately model attraction~\cite{erlander1990gravity}, but still fall short of properly modeling complex nonlinearities, such as interactions between urban utilities and human mobility.
More recent approaches have used the radiation model~\cite{simini_universal_2012,yan_universal_2014}, derived from a stochastic process considering intervening opportunities. Radiation models are limited by data capacity, using only population distribution and ignoring the growing availability of urban data~\cite{doi:10.1089/big.2014.0020}.
Machine learning approaches have also been proposed for trip distribution modeling, including random forest~\cite{robinson_machine_2018, pourebrahim_enhancing_2018,pourebrahim_trip_2019}. These machine learning models make use of rich urban data and can better model complex nonlinearities.
However, these models ignore the spatial correlations and consider only the characteristics of origin and destination.
In our previous work~\cite{liu2020gmel}, we proposed to use a graph neural network to learn geo-contextual embeddings for commuting trip distribution modeling, achieving better predictive performance when compared against baseline models, such as gradient boosting regression tree, random forest, gravity model, and node2vec. 

\subsection{Graph representation learning}
Graph representation learning aims at learning low-dimensional features (i.e., embeddings) for each node in a graph, preserving both the graph structure and node attributes. This approach allows the embeddings to be used in a myriad of analytical tasks, such as community detection~\cite{li2018community}, traffic prediction~~\cite{Wang:2017:RRL:3132847.3133006} and graph isomorphism~\cite{xu2019powerful}.
Graph neural networks~(GNN) provide powerful graph embedding capabilities~\cite{xu2019powerful}. Different approaches include graph convolution neural network~(GCN), based on the notion of convolution on graphs \cite{bruna2014spectral}, a general inductive framework that leverages node attributes to generate node embeddings in a message-passing way~\cite{hamilton_inductive_2017}, and graph attention networks that leverage self-attention mechanisms to allow messages passed by neighbors to be aggregated with different weights~\cite{velickovic_graph_2017}.
Successful applications include traffic prediction~\cite{yu2020forecasting}, recommender system~\cite{ying2018graph}, and drug discovery~\cite{kearnes2016molecular}.
In this work, we use our previous work on graph attention networks with a modified attention mechanism so that the model can capture the spatial correlations and enable the construction of alternate scenarios~\cite{liu2020gmel}.

\subsection{Scenario planning}
The ability to plan for different scenarios and consider different outcomes is important in several domains, including urban planning and transportation. At its core, scenario planning allows planners to analyze future outcomes based on present-day decisions~\cite{doi:10.1080/01944363.2015.1038576}.
Traditional approaches usually use regression analysis~\cite{doi:10.1080/01944363.2011.582394}, travel forecasting models, or econometric  models~\cite{bartholomew2007land}.
More recently, machine learning approaches have been proposed to best guide stakeholders on how to best plan for future growth, while taking into account environmental considerations~\cite{KIM2020101498}, or how to best calculate land use configuration given surround spatial contexts~\cite{wang2020reimagining}.
Simultaneously, given that transparency, expert feedback and community participation is an increasingly important topic in scenario planning, different proposals have considered a human-in-the-loop approach to foster the involvement of stakeholders~\cite{8283638, ortner2016vis, Ferreira2015Urbane:Development, Doraiswamy:2018:IVE:3183713.3193559}.
\section{Methodology}
\label{sec:methodology}

In this work, we leverage our previously proposed graph neural network model, called Geo-contextual Multitask Embedding Learner (GMEL)~\cite{liu2020gmel}. Next, we briefly describe GMEL and also how we use the model for scenario planning in two cities in Brazil.

\subsection{GMEL}
GMEL is a graph neural network model for commuting trip distribution modeling. The model consists of two components: a geo-contextual multitask embedding learner and a flow predictor.
The learner was designed to capture the spatial correlations from geographic neighborhoods. The model utilizes a graph attention network (GAT) to encode the spatial dependencies into an embedding space. To disentangle the origin and destination characteristics that are hidden in the infrastructure and land use data, GMEL employs two separate GATs to encode the geographic contextual information into two different embedding spaces.
GMEL employs multitask learning framework which imposes stronger restrictions forcing the embeddings to encapsulate effective representations for flow prediction.
The second component, the flow predictor, employs a gradient boosting machine (GBM) as the regression model to predict commuting flows. GBM iteratively evaluates the largest information gain of features, automatically selecting and combining useful numerical features to fit the targets.

\begin{figure}[t]
\centering
\includegraphics[width=1.0\linewidth]{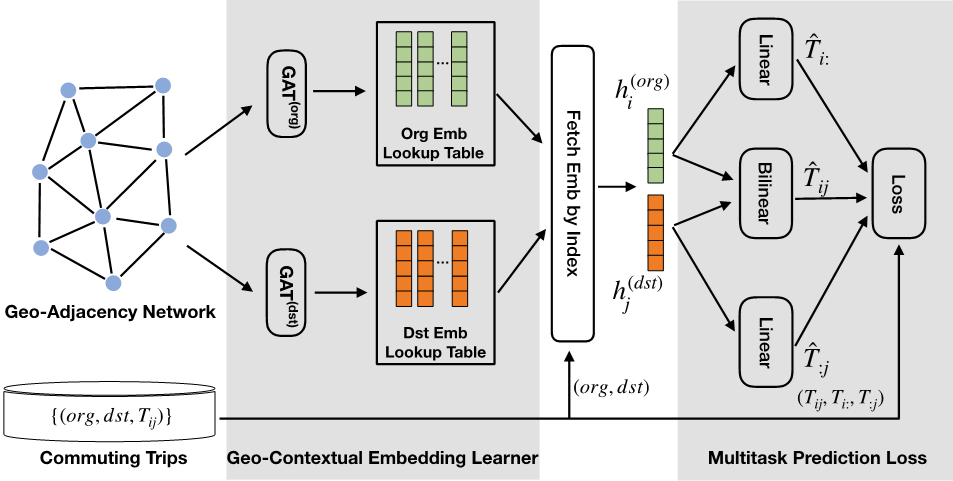}
\caption{GMEL architecture~\cite{liu2020gmel}.}
\label{fig:overview}
% \vspace{-0.8cm}
\end{figure}

Figure~\ref{fig:overview} presents an overview of the GMEL architecture. The model makes use of two datasets: 
1) a set of origin-destination commuting trips, where each trip is composed of an origin, a destination, and a number of commuters traveling from origin to destination. Origin and destinations are aggregated at a census tract level, and commuting flows are yearly static values, i.e., the number of people that reports living in an origin census tract and working at a destination census tract.
2) a geo-adjacency network, an undirected weighted graph with the following properties: geographic units (i.e., census tracts) as nodes, with an associated set of urban indicators, and the weight of edges describe the strength of correlations between units (i.e., travel distance, trip duration). 

GMEL was originally proposed using an extensive land use dataset from New York City, called Property Land Use Tax Lot Output (PLUTO). The dataset describes, for each year between 2008 and 2021, the land use information of the city, at a lot level.
This enabled us to perform initial \emph{what-if} scenario explorations. In one specific case study, we showed how we could leverage a GMEL model that was trained using the PLUTO data set for one specific year (2013) and predict flows considering the modified land use of a subsequent year (2015), highlighting how GMEL can guide urban planners and policymakers to make informed decisions when it comes to new urban development scenarios.

% \begin{table*}[h!]
% \centering
% \caption{Summary of urban indicators.}
% \begin{tabular}{ccccc}
% \hline
% & \multicolumn{2}{c}{Recife} & \multicolumn{2}{c}{Curitiba}\\
%  Category & \# Features & Content & \# Features & Content \\ 
% \hline
% \hline
% Infrastructure & 30 & Number of different types of buildings (12), density of residential/etc. units (4), number of buildings in each built year interval (11), the number of bike stations (1), the perimeter of bikelanes (1), the perimeter of bus lanes (1). & 12 & The perimeter of bikelanes (1), number of lots per zone (11).\\

% Land use & 14 & Land area ratio of retail/office/etc. (12), statistics of floor area ratio (2). & 11 & Land area ratio of zones (11).\\ 

% Speciality & 2 & Whether or not the urban geographic unit contains landmarks (1), number of cultural routes (1) & 1 & Number of lots contained in a historic area (1).\\
% \hline
% Total & &\\
% \hline
% \end{tabular}
% \label{table:indicators}
% \end{table*}

\begin{table*}[t!]

\footnotesize
    \centering
    \caption{Summary of urban indicators for Recife and Curitiba.}
    \vspace{-0.3cm}
      \begin{tabular}{ccccc}
      \toprule
      \multirow{2}[4]{*}{\textbf{Category}} & \multicolumn{2}{c}{\textbf{Recife}} & \multicolumn{2}{c}{\textbf{Curitiba}} \\
  \cmidrule{2-5}    \multicolumn{1}{c}{} & \multicolumn{1}{p{.05\linewidth}}{\textbf{\# Feat.}} & \multicolumn{1}{c}{\textbf{Content}} & \multicolumn{1}{p{.05\linewidth}}{\textbf{\# Feat.}} & \multicolumn{1}{c}{\textbf{Content}} \\
      \midrule
      \multirow{4}[0]{*}{Infrastructure} & \multirow{4}[0]{*}{30} & \multicolumn{1}{p{.33\linewidth}}{No. of different types of buildings (12), density of res. units (4), no. of buildings per built year (11), no. of bike stations (1), perimeter of bike lanes (1), perimeter of bus lanes (1)} & \multirow{4}[0]{*}{12} & \multicolumn{1}{p{.33\linewidth}}{Perimeter of bike lanes (1), no. of lots per zone (11)} \\
      \midrule
      \multirow{1}[0]{*}{Land use} & \multirow{1}[0]{*}{14}    & \multicolumn{1}{p{.33\linewidth}}{Land area ratio of retail/office (12), floor area ratio (2)} & \multirow{1}[0]{*}{11} & \multirow{1}[0]{*}{Land area ratio of zones (11)} \\
      \midrule
      \multirow{2}[0]{*}{Speciality} & \multirow{2}[0]{*}{2}    & \multicolumn{1}{p{.33\linewidth}}{Whether or not the urban geographic unit contains landmarks (1), no. of cultural routes (1)} & \multirow{2}[0]{*}{1}     & \multirow{2}[0]{*}{Number of lots contained in a historic area (1)} \\
    %   \midrule
    %   Total & 46    &       & 24    &  \\
      \bottomrule
      \end{tabular}%
    \label{table:indicators}
  \end{table*}%

\subsection{Scenario planning with GMEL}
In this paper, we further explore the possibilities of using GMEL for \emph{what-if} scenario planning. We focus our efforts on two Brazilian cities: Recife, the oldest capital city of Brazil, the 3rd most populous city in the Northeast region of Brazil, and the 9th most populous in the country;
Curitiba is the most populous city in the South region of Brazil, and the 8th most populous city in the country.
Curitiba in particular is known for its innovative urban planning initiatives, targeted at improving public transportation accessibility and promote housing development.

Recife and Curitiba are among the fastest-growing cities in Brazil~\cite{topbrazilcities}, creating the need to have the right methods in place to allow stakeholders to better plan urban interventions. This need was recently highlighted by a report from the city of Recife, that states the \emph{``need to develop analytical tools that allow the creation of scenarios to capture specific changes in certain areas of the city, and to allow the assessment of scenarios related to the implementation of new transportation infrastructure.''}
At the core of our proposal is the ability to use a graph neural network (i.e., GMEL) to predict changes in commuting flows, given changes in the land use and built environment. We follow a set of steps that allow us to train and validate a model, and then use this model to predict flows in a different scenario. Our steps can be summarized as follows:

\myparagraph{Model training:} we initially train a GMEL model for each city, considering land use and commuting flow data from Recife and Curitiba. 
The model is trained using stochastic gradient descent in an end-to-end manner. With the embeddings from the trained GMEL, a GBRT is trained as a flow predictor based on the concatenation of origin-destination embeddings and travel distances to predict the commuting flow.
The model is then tested using a holdout set.

\myparagraph{Scenario planning:} after the model is trained and tested, we change the urban indicators of the city, following plausible urban modification scenarios currently being proposed in the two cities.
We update the geo-adjacency network to follow these modifications and use the previously trained models to generate \emph{new} embeddings for the modified network.
The flow prediction then uses these embeddings to predict new commuting flows.

\section{Model Training and Results}
\label{sec:results}

This section describes the input generation process for the GMEL model and the experimental results for both Recife and Curitiba.

\subsection{Data description}
\label{sec:description}

To train and validate the models used for scenario planning, we used open datasets from Recife and Curitiba. In both cases, we used the cities' 2020 census tracts as the \emph{geographic units}. To measure the travel distance between (the centroids of) the census tracts, the Open Source Routing Machine (OSRM) was used, as described in~\cite{liu2020gmel}.
Next, we detail the datasets and their sources. The information is also summarized in Table~\ref{table:indicators}.

The first city studied in this paper is the city of Recife, capital of the state of Pernambuco in the Northeast coast of Brazil. For this analysis, we used the commuting flow dataset obtained from a survey carried out by the city's administration between 2018 and 2019. This survey captured data on typical commuting movements performed by the population that resides, works, studies, or seeks services in the metropolitan region of the city. 
As urban indicators, we used information about individual lots, as well as indicators that show the presence of special preservation buildings, cultural routes, bicycle stations, bike lanes, and exclusive bus lanes, all available on Recife's open data portal \cite{dadosabertos}.
After joining flow and the indicators datasets, we ended up with 1,347 census tracts that cover more than half of the city. The commuting flows were aggregated into geographic unit level flows, resulting in 23,336 commuters and 15,945 pairs of origin-destination trips divided into train (60\%), test (20\%), and validation (20\%) datasets.
% The few tracts with missing data for at least one of the datasets were removed.

The second city considered is Curitiba, the capital of the state of Paran\'{a} in the South region of Brazil. The commuting flow dataset used for the analysis was obtained from a survey carried out by the city's administration in 2017, mapping commuting patterns in the metropolitan region of the city.
As urban indicators, we used individual lots aggregated by zone, according to the new zoning legislation, established in 2020, as well as the distribution of bike lanes in the city, data made available by Curitiba's urban planning and research institute~\cite{ippuc}.
After joining commuting flows and indicators, we ended up with 2,200 census tracts, covering the vast majority of the city. The commuting data resulted in 45,365 commuters and 32,988 pairs of origin-destination trips, divided in the same way as previously described.

In our case studies, we modify the previously mentioned urban indicators to effectively predict \emph{new} flows when considering alternate scenario plans.

\subsection{Performance analysis}
\label{sec:performance}

We evaluated the performance of the GMEL model in both Recife and Curitiba.
To measure the prediction performance, we adopted three evaluation metrics: Root Mean Square Error~(RMSE), Mean Absolute Error~(MAE), and Common Part of Commuters~(CPC). For each pair $i,j$ of nodes, and predicted flow value $\hat{T}_{i,j}$ and groundtruth flow value $T_{i,j}$, we have $\mathrm{RMSE} = \sqrt{ \dfrac{1}{|T|} \sum_{i,j} (\hat{T}_{ij} - T_{ij})^2 }$, $\mathrm{MAE} = \dfrac{1}{|T|} \sum_{i,j} |\hat{T}_{ij} - T_{ij}|$, and $\mathrm{CPC} = \dfrac{2\sum_{ij}min(\hat{T}_{ij}, T_{ij})}{\sum_{ij}\hat{T}_{ij} + \sum_{ij}T_{ij}}$.

% between nodes $i,j$, and by $T_{i,j}$ the groundtruth flow between nodes $i,j$.

% \begin{align}
%     \mathrm{RMSE} &= \sqrt{ \dfrac{1}{|T|} \sum_{i,j} (\hat{T}_{ij} - T_{ij})^2 } \\
%     \mathrm{MAE} &= \dfrac{1}{|T|} \sum_{i,j} |\hat{T}_{ij} - T_{ij}| \\
%     \mathrm{CPC} &= \dfrac{2\sum_{ij}min(\hat{T}_{ij}, T_{ij})}{\sum_{ij}\hat{T}_{ij} + \sum_{ij}T_{ij}}
% \end{align}
% \noindent where $\hat{T}_{i,j}$ is the predicted flow between nodes $i,j$, and $T_{i,j}$ is the groundtruth flow between nodes $i,j$.

RMSE and MAE are widely used as evaluation metrics for regression problems. CPC is widely used in trip distribution modeling~\cite{lenormand_systematic_2016,robinson_machine_2018}, and it measures the agreement between the predicted value and target value; CPC is 0 when no agreement is found, and it is 1 when the two are identical.

% \begin{table}[h]
% \centering
% \caption{Performance on test set.}
% \vspace{-0.3cm}
% \begin{tabular}{cccc}
% \hline
%  City & RMSE & MAE & CPC* \\ 
% \hline
% \hline
% Recife & 1.43 & 0.73 & 0.73 \\
% \hline
% Curitiba & 1.08 & 0.70 & 0.73\\
% \hline
% \multicolumn{4}{l}{* Higher is better}
% \end{tabular}
% \label{table:results}
% \end{table}

The model's performance for the two cities was quite similar (Table~\ref{table:results}), with the RMSE being the metric with the most significant difference, although still small. This can be attributed to the difference in the number of commuting flows and census tracts used in Curitiba and Recife. Given that this number was greater in Curitiba, a lower RMSE was expected.

\begin{table}[h!]
\centering
\caption{Performance on test set.}
\vspace{-0.3cm}
\begin{tabular}{cccc}
\hline
 City & RMSE & MAE & CPC* \\ 
\hline
\hline
Recife & 1.43 & 0.73 & 0.73 \\
\hline
Curitiba & 1.08 & 0.70 & 0.73\\
\hline
\multicolumn{4}{l}{* Higher is better}
\end{tabular}
\label{table:results}
\vspace{-0.5cm}
\end{table}
\section{Case Studies}
\label{sec:cases}

% We now describe examples of alternate planning scenarios generate by our models.
Following our methodology, we now present two initial alternate scenario plans for the city of Recife. Both of the case studies highlight the importance of developing tools and frameworks that allow the creation of scenarios to assess the impact of land use changes.

\subsection{New bike lanes in Recife}
\label{sec:bike_use_case}

\begin{figure}[t]
\centering
\includegraphics[width=0.95\linewidth]{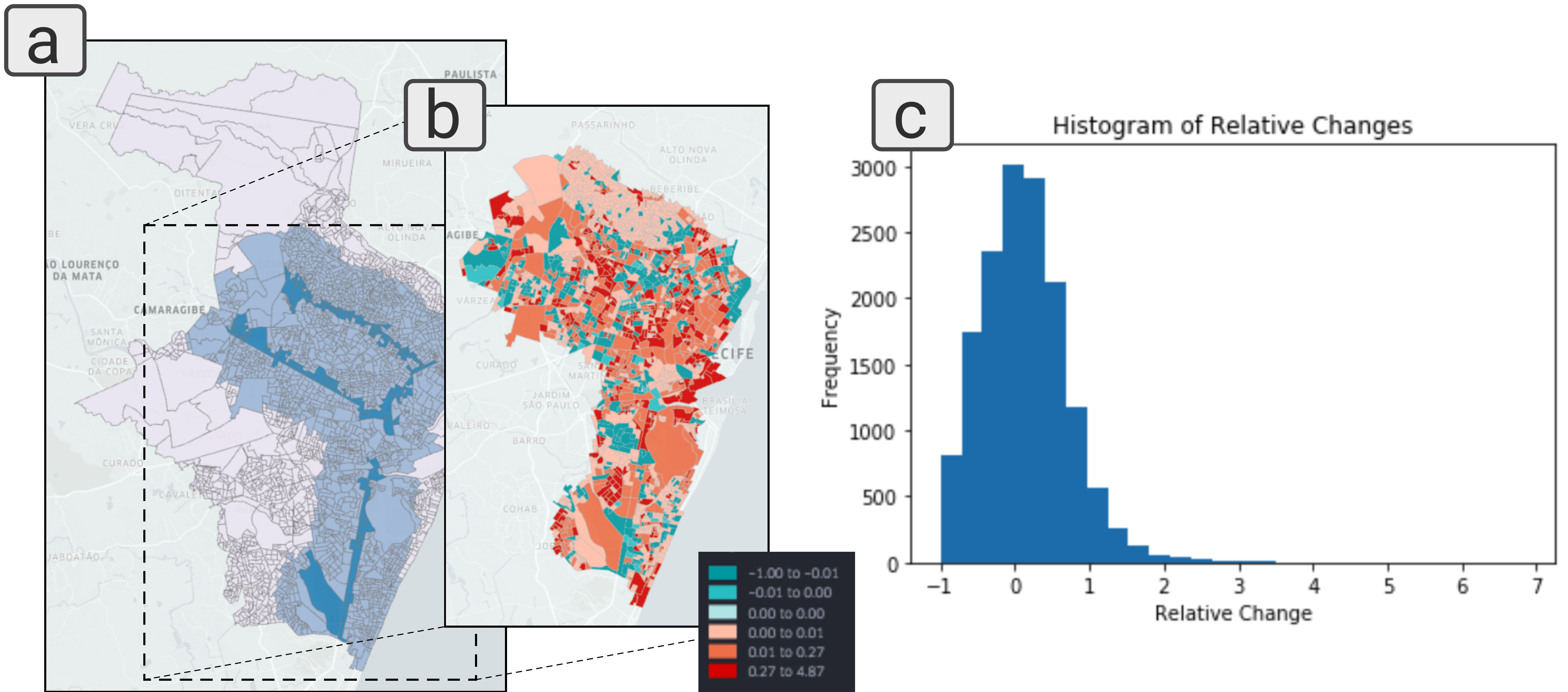}
\caption{Alternate land-use scenario considering new bike lanes: (a) census tracts new bike lanes (dark blue), and tracts with centroids within 2~km of distance (light blue); (b) commuting flow differences, considering the new scenario; (c) histogram with commuting flow differences.}
\label{fig:bikes}
\end{figure}

In recent years, there has been a growing movement in which citizens organize themselves to think about alternatives that go beyond motor vehicles. 
Bicycles are means of transport for small and medium distances that bring benefits to health and the environment.
In some cities around the world, bicycles are already a widely used mode of transportation for a significant portion of the population, while in Recife it is mostly used by a small portion of the population. 
While there has been an increasing demand for new cycle paths around the city, an important question that should be answered before actually creating these new paths is whether this infrastructure will effectively generate an increase in the flow of cyclists in the region. 
To show how to apply our methodology in this scenario, we chose four major avenues that connect important points in Recife.
% (Av. Governador Agamenon Magalh\~{a}es, Av. Norte Miguel Arraes de Alencar, Av. Mascarenhas de Moraes and Av. Caxang\'{a}). 
%
These avenues have a high rate of daily vehicle flow, however, currently they do not have any type of cycling infrastructure, posing a risk to cyclists who need to travel around them to reach their destination. We distributed 24~km of cycle paths between these four avenues, mapped to the zones where the avenues are located (Figure~\ref{fig:bikes}(a)). 
To analyze the changes caused by these modifications we looked at the flows among units whose geographical centers are up to 2~km away from the centers of units that received the new cycle paths.
For each flow, we computed the relative change as a way to assess the amount of flow change in these areas.
The results show an average increase of 13\% (and 0.59 std. deviation) of flows among those units (Figure~\ref{fig:bikes}(b,c)).
While our model did not use any data on the modality of transportation, we hypothesize that these changes could represent an increase in biking in these areas.

\subsection{New high-rise buildings in Recife}

\begin{figure}[t]
\centering
\includegraphics[width=0.95\linewidth]{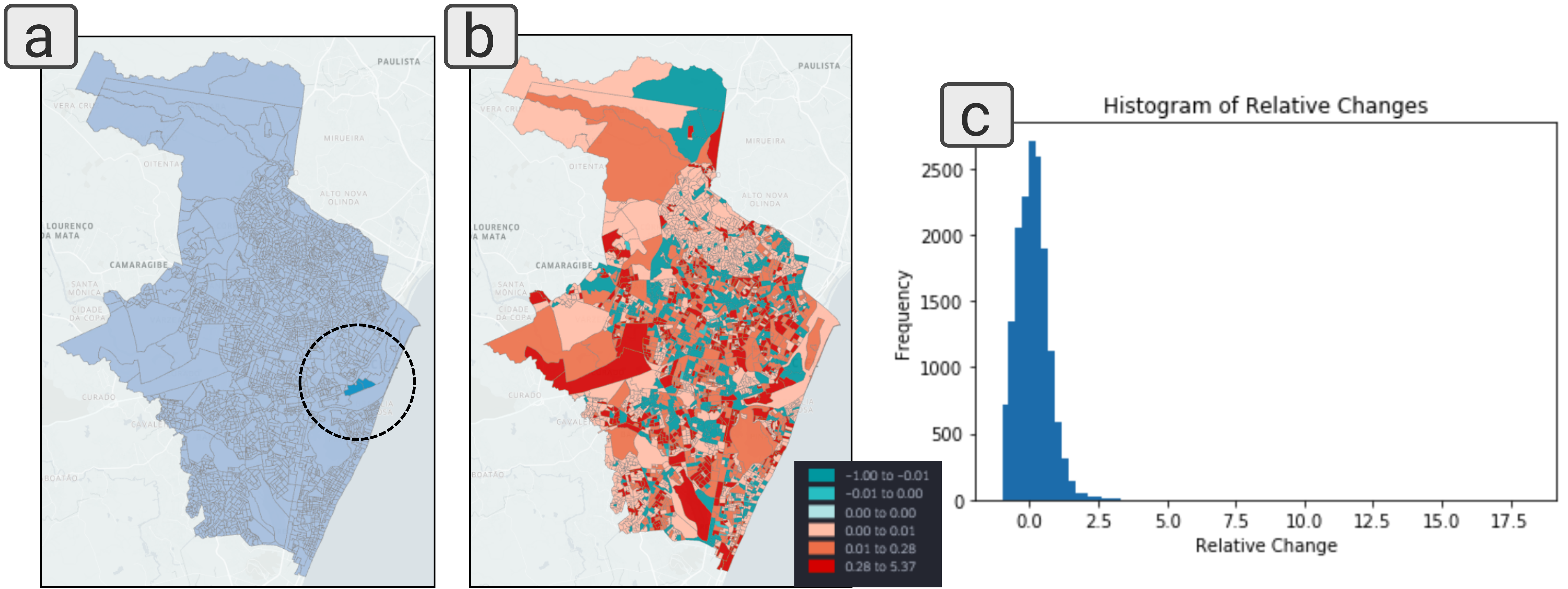}
\caption{Alternate land-use scenario considering new buildings: (a) census tracts with new buildings (dark blue); (b) commuting flow differences, considering the new scenario; (c) histogram with commuting flow differences.}
\label{fig:buildings}
\vspace{-0.5cm}
\end{figure}

Building large projects in a city is an activity that is always accompanied by a series of impacts, whether on the landscape, on the region's economy, and/or on the flow of people moving from one point to another. 
It is therefore of great importance to urban planners to first estimate these impacts on the city's infrastructure in order to anticipate actions to fulfill new needs or solve possible upcoming problems.
As a way to estimate the impact on the flow of people due to a large construction in downtown Recife, we used the real-world case of the construction of 13 towers, with sizes ranging from 13 to 44 floors, on the former historical site of the Jos\'{e} Estelita Pier (Figure~\ref{fig:buildings}(a)). We incorporated the new buildings into the model, adding information related to the number of different types of buildings, density, built year, land/area ratio, and floor/area ratio.
Similar to Section~\ref{sec:bike_use_case}, we analyzed the relative changes in commuting flows that were caused by the modification of the urban indicators.
%
% We performed a similar analysis as in Section~\ref{sec:bike_use_case} to analyze relative changes in commuting flows due to these urban indicator modifications.
%
The results show an average increase of 12,5\% (0.605 std. deviation) of flows among those units (Figure~\ref{fig:buildings}(b,c)).
Notice that the model predicts relative changes thought the city. However, the most significant changes happen in units close to the new project, or in high-density units that already have large commuting flow volumes.
% is locate ora few distant populous ones for which large intensity commuting flows already exist. 
These flows are predicted to increase by at least 25\%.

\section{Conclusions}
\label{sec:conclusion}
Our goal in this paper was to analyze the behavior of flows on proposed \emph{what-if} scenarios using the GMEL model, evaluating the model's performance with available urban data from two different cities in Brazil.
Our initial results show that the model could be used as a tool to assist urban planners and transportation researchers in the decision-making process, with good performance even when considering cities in developing countries. There are, however, challenges regarding data quality that need to be addressed.
Similar to the majority of data-driven approaches, GMEL is heavily dependent on the quality of the datasets used. It is then important to have an effective involvement of public administration in providing accurate information to be used in the model.
%
% This paper presents an initial exploration of the capabilities of graph neural networks approaches for scenario planning. 
%
In future work, we plan to further evaluate our model and extend it to take into account multimodal commuting flows, which will allow us to answer more specific questions regarding people's movement in the city considering different modes of transportation.

% The results show a promising results of the model for the two cities studied as well as interesting patterns for spatial units that  have undergone some change. GMEL is a model with potential to assist urban planners in the decision-making process, but there are challenges to be overcome regarding the maturity of the datasets used, which can significantly interfere in the accuracy of the predictions made, and therefore it is necessary an effective involvement of the public administration in providing the information that will feed the model. One way for future work is to use GMEL in multimodal commuting flows datasets, which should allow answering more specific questions about the movement of people in the city by type of transport used.

\bibliographystyle{ACM-Reference-Format}
\bibliography{main}

\end{document}